\documentclass[runningheads]{llncs}

\usepackage{times}
\usepackage{latexsym}
\usepackage{amsmath}
\usepackage{multirow}
\usepackage{url}
\usepackage{dcolumn}
\usepackage{enumitem}
\usepackage{longtable}
\usepackage{color, colortbl}
\definecolor{Gray}{gray}{0.95} 
\usepackage{subcaption}
\usepackage{adjustbox}
\usepackage{tabularx}
\usepackage[T1]{fontenc}

\usepackage[T1]{fontenc}

\usepackage{microtype}
\usepackage{dirtytalk}

\usepackage{enumitem}
\newlist{inlinelistone}{enumerate*}{1}
\setlist*[inlinelistone,1]{%
  label=(\textbf{\arabic*)},
}
\newlist{inlinelisttwo}{enumerate*}{1}
\setlist*[inlinelisttwo,1]{%
  label=(\textbf{\roman*)},
}

\usepackage{inconsolata}
\usepackage{graphicx}
\usepackage{comment}

\usepackage{color}

\title{Continual Learning Under Language Shift}

\author{Evangelia Gogoulou\inst{1,2} \and
Timothée Lesort\inst{3,4} \and
Magnus Boman\inst{5,6} \and
Joakim Nivre\inst{1,7}
}

\authorrunning{E. Gogoulou et al.}
\institute{RISE Research Institutes of Sweden \and KTH Royal Institute of Technology \and Université de Montréal \and MILA-Quebec AI Institute \and
Karolinska Institutet \and MedTechLabs \and Uppsala University \\
Correspondence: \email{evangelia.gogoulou@ri.se}\\}

\begin{document}
\maketitle
\begin{abstract}


The recent increase in data and model scale for language model pre-training has led to huge training costs. In scenarios where new data become available over time, updating a model instead of fully retraining it would therefore provide significant gains.
We study the pros and cons of updating a language model when new data comes from new languages -- the case of continual learning under language shift.
Starting from a monolingual English language model, we incrementally add data from Danish, Icelandic and Norwegian to investigate how forward and backward transfer effects depend on pre-training order and characteristics of languages, for models with $126$M, $356$M and $1.3$B parameters. 
Our results show that, while forward transfer is largely positive and independent of language order, backward transfer can be 
positive or negative depending on the order and characteristics of new languages. 
We explore a number of potentially explanatory factors and find that a combination of language contamination and syntactic similarity 
best fits our results. 

\end{abstract}


\section{Introduction}
\label{sec:intro}

Pre-training on large amounts of text is the current paradigm for training language models. 
While 
these models 
have good generalization capabilities for 
downstream tasks, they will 
soon become obsolete given the current rate of producing new data that differ from the training data 
\cite{Lazaridou2021Mind}. Continual learning 
aims to update a model with new data while retaining the previous model knowledge in different scenarios of data distribution shift \cite{Parisi18review}. 
In this paper, we focus on the specific case of language shift, where each new data source is in a different language, and study how the model learns and forgets when sequentially trained on multiple languages, 
in a standard language modeling scenario.
From a continual learning perspective, this multilingual scenario is particularly 
interesting because the distribution shift 
between two languages is expected to have a big impact on model knowledge. In addition, 
we can study cross-lingual transfer effects as a monolingual model becomes progressively multilingual.

The main research question of this paper is: How do the characteristics of languages and their order in the task sequence influence model performance? We focus on how the model performs on new languages -- \emph{forward transfer} -- and on previous languages -- \emph{backward transfer}. We do this by training generative language models on data from English, Danish, Icelandic, and Norwegian, in different sequential
orders. We also investigate whether increasing the model capacity alleviates catastrophic forgetting of previous languages or changes the transfer effects in other respects. 
Finally, in an attempt to explain the transfer effect patterns, we explore 
language similarity metrics 
and track language contamination in training data.

Our main findings are: \begin{inlinelisttwo}
    \item 
    Forward transfer 
    is consistently positive and language-independent.
    
    \item 
    Backward transfer can be positive or negative depending on the specific languages and the order in which they are added. 

    \item 
    Increasing model size improves final performance on all languages but does not 
    affect transfer patterns. 

    \item 
    Backward transfer effects 
    correlate with 
    language contamination as well as syntactic similarity.
    \end{inlinelisttwo}

\section{Related Work}
\label{sec:related}
%
%
 \textit{Continual Learning}
Continual learning in the deep 
learning literature generally addresses the problem of learning without forgetting on sequences of datasets \cite{French99}. 
The 
classic setup for 
this problem is image classification \cite{NEURIPS2019_15825aee,douillard2021dytox}. 
However, recent progress has led to increased interest in continual learning also in NLP \cite{Autume2019Episodic,biesialska-etal-2020-continual}. 
Notably, the increased availability of large pre-trained models 
provides good foundations for continual learning where only slight adaptations are needed for downstream tasks 
\cite{abnar2021exploring}. 
We study the influence of model size on continual performance as in previous studies \cite{ramasesh2022effect,Lesort2022Challenging}. 

\textit{Continual Language Pre-Training}
Continual pre-training of language models has been previously applied to data evolving through time \cite{han2021econet,jang-etal-2022-temporalwiki,loureiro2022timelms}. 
In our setup, we assume 
factual knowledge to be invariant and are interested in incorporating more 
knowledge while minimizing forgetting as in domain adaptation \cite{ke2023continual,scialom2022fine}. 
As the capacity of algorithms to continuously learn 
depends on 
data distribution shifts \cite{lesort2021understanding}, we study the 
multilingual setup as a corner case with 
drastic distribution changes. 

\textit{Multilingual Language Models}
Examples of multilingual Transformer encoder models are mBERT \cite{devlin-etal-2019-bert} and XLM-R \cite{conneau2020unsupervised}. 
Following the development of Transformer decoder architectures, several multilingual GPT-style \cite{radford2018improving} models have been proposed, such as BLOOM \cite{scao2022bloom}, 
XGLM \cite{lin-etal-2022-shot}, Mixtral \cite{jiang2024mixtral} and GPT-SW3 \cite{ekgren2023gpt}. 
Joint pre-training is used as a baseline in our setup, while our work focuses on multilingual models where each new language is added to the model one by one, starting from scratch. 
Our model architecture, data, and training process closely follow GPT-SW3 \cite{ekgren2023gpt}. 
Studies related to continual learning of multilingual models include a method for continual fine-tuning of mBERT on unbalanced language distributions \cite{badola2022parameter}, 
 a study of catastrophic forgetting when continually fine-tuning mBERT on two sequence labeling tasks one language at a time \cite{coria-etal-2022-analyzing},  
a novel continual multilingual fine-tuning method, 
which opts for both minimum forgetting in the source languages and reduction of the fine-tuning loss \cite{Yuren2022Less},
and a study of continual pre-training of language models, focusing on improving algorithmic solutions \cite{winata2023overcoming}. 
It has also been shown that unexpected multilingual capabilities of monolingual models can be explained by 
language contamination: data from other languages accidentally being included in monolingual training data \cite{blevins-zettlemoyer-2022-language}.

\section{Method}
\label{sec:method}

\subsection{Learning Scenario and Setup}


We consider a scenario where a generative language model is trained in successive stages given a data input stream $D={D_{1},D_{2}, \ldots D_{n}}$, 
where
each $D_{i}$ contains data from a different language. After each stage, we evaluate model performance on each language seen during training by computing cross-entropy loss on held-out test sets. To study how forward and backward transfer effects depend on the order and characteristics of languages involved, we systematically vary the order of languages.

In this study, we experiment with data from four languages, namely English (en), Danish (da), Icelandic (is), and Norwegian (no). We consistently take the initial data $D_1$ from English, and then add Danish, Icelandic and/or Norwegian in all possible orders, which gives us 15 different training sequences of different lengths.\footnote{3 sequences of length 2 (English + 1 language) and 6 sequences each of length 3 and 4.}All models are randomly initialized before being trained on the first language of the sequence. For each successive continual learning stage $i$ ($i > 1$), all model parameters are initialized from the last checkpoint obtained after stage $i-1$ and then updated using the standard next-step prediction objective with data batch $D_{i}$ as input. 

The size of the training sets is constant across languages, which allows us to study the effect of language shift independently from data set size. Regarding tokenization, we assume that a multilingual tokenizer, covering all the languages studied, is available to us before training. This is often the practice case as well, given the abundance of openly available tokenizers with large language coverage \cite{conneau2020unsupervised,scao2022bloom}, although it is an unrealistic assumption for many low-resource languages. 
For comparison, we also train two types of baseline models: one monolingual model for each language (en, da, is, no), trained on the corresponding training batch $D_i$, and one multilingual model (en+da+is+no) trained on the concatenation of the three training sets $D_1+D_2+D_3$. 

\subsection{Model Architecture and Training}
We use the standard GPT architecture \cite{radford2018improving}, as implemented in the Nvidia NeMo library.\footnote{\url{https://github.com/NVIDIA/NeMo}} To study the role of model capacity, we run experiments with three model sizes: $126$M,  $356$M, and $1.3$B parameters. Exact model hyperparameter values, following \cite{ekgren2023gpt}, are given in Table \ref{tab:model_hyperpar} in Appendix~\ref{app:details}. 
For tokenization, all models use the GPT-SW3 tokenizer \cite{ekgren2023gpt,stollenwerk2023training}, which is a BPE tokenizer with $64$K tokens, trained on the Nordic Pile \cite{ohman2023nordic}, a multilingual corpus including English, Danish, Icelandic, Norwegian, and Swedish. For each language pre-training stage during continual learning, the model is trained for $35$K steps (approximately $36$M sequences of $2048$ tokens each). The monolingual baselines are also trained for $35$K steps, while the multilingual baseline model is trained for $105$K steps, thereby covering the training sets for all three languages. 
Following \cite{ekgren2023gpt}, we start by warming up the learning rate for $250$ steps, increasing its value to the maximum, and then decreasing it according to the cosine decay function until reaching its minimum value, where it will stay constant for the last $4.9$K steps. 

\subsection{Datasets}

Models are trained on a subset of the Nordic Pile corpus \cite{ohman2023nordic}.\footnote{See Appendix \ref{app:details} for more information about the total corpus and data weighting scheme.} 
The list of datasets used per language, together with the number of training sequences per dataset is presented in Table~\ref{tab:lang_corpora}. Each of the validation and test sets per language consists of $51$K samples and has the same proportions of datasets as the training set.

\begin{table}[t]

\centering
\caption{Data per language (Lang) and number of training sequences in millions (N).}
\label{tab:lang_corpora}
\vspace{1mm}
\renewcommand{\tabcolsep}{3pt}
\scalebox{0.87}{
\begin{tabular}{|l|r||l|r||l|r||l|r|}
\hline
\textbf{English} & \textbf{N} & \textbf{Danish} & \textbf{N} & \textbf{Icelandic} & \textbf{N} & \textbf{Norwegian} & \textbf{N} \\
\hline
The Pile \cite{gao2020pile} & $32$ & Wiki da & $0.36$ & Icelandic Gigaword \cite{barkarson2022evolving} & $24$ & NCC \cite{kummervold-etal-2022-norwegian} & $18$ \\

Wiki en & $4$ & mc4 (da) \cite{xue-etal-2021-mt5} & $18$ & Wiki is & $0.23$ & Wiki no & $0.37$ \\

& & OSCAR (da) \cite{ortiz-suarez-etal-2020-monolingual} & $16$ & mc4 (is) \cite{xue-etal-2021-mt5} & $6$ & mc4 (no) \cite{xue-etal-2021-mt5} & $11$ \\

& & Danish Gigaword \cite{stromberg-derczynski-etal-2021-danish} & $0.36$ & OSCAR (is) \cite{ortiz-suarez-etal-2020-monolingual} & $6$ & OSCAR (no) \cite{ortiz-suarez-etal-2020-monolingual} & $7$ \\
\hline
\end{tabular}
}
\end{table}

\subsection{Language Similarity Metrics}
\label{sec:sim}
One of our hypotheses is that the degree of shift between language corpora 
will have a significant effect on the model knowledge 
and hence we explore two types of metrics for estimating the similarity between languages and their distributions.
First, the linguistic similarity between two languages is used, as estimated by the pre-computed distance\footnote{\url{http://www.cs.cmu.edu/~dmortens/projects/7_project/}} of their corresponding \textit{lang2vec} language vectors which are part of the URIEL database \cite{littell-etal-2017-uriel}. Different types of features are considered: syntactic (SYN), 
phonetic (INV), and phonological (PHON).\footnote{While phonetic and phonological features are not directly relevant for written texts, they can be seen as proxies for orthographical features.}
%
%
Second, we introduce a data-driven metric of similarity between two language corpora, which we call \emph{token distribution similarity} (TDS). The idea is 
that similarity should take into account the difference in token distributions between two datasets. Hence, we create one vector per dataset, counting the number of occurrences for each token, and then compute the cosine similarity between vectors of different datasets. 
This is similar to the metric of \cite{garcia2021continual}, which estimates the similarity between two vocabularies by the percentage of token overlap between them. The advantage of our method is that we do not take into account only the list of common tokens, but also the frequency of their occurrence.
We sample $50$K sequences for each language, according to the data weighting scheme used for pre-training. We tokenize each language sample using the model tokenizer and create a vector where position $i$ corresponds to the frequency of vocabulary token $i$ in the sample. Next, we compute the cosine similarity of the two token frequency vectors, which is used as a proxy for language similarity. 

The similarity values between all language pairs 
are shown in Table \ref{tab:language_sim}. 
According to the TDS metric, Danish and Norwegian have by far the highest degree of similarity, followed by English-Norwegian and English-Danish, while English and Icelandic have the lowest similarity. For the URIEL features, 
we see very different patterns for syntactic features, where English, Danish and Norwegian are most similar while Icelandic is an outlier, and for phonetic and phonological features, where the Scandinavian languages are highly similar and different from English.
Our hypothesis is that positive transfer will be observed for languages with the highest TDS value, as this metric derives from the overlap between the data inputs, but we also expect syntactic similarity to be relevant. 
\begin{table}[t]
\centering
\caption{Language similarity values 
estimated by our 
TDS metric and by URIEL 
syntactic (SYN), phonetic (INV) and phonological (PHON) features \cite{littell-etal-2017-uriel}.}
\label{tab:language_sim}
\vspace{1mm}
\begin{adjustbox}{width=0.8\textwidth}
\begin{tabular}{|l|D{.}{.}{2}|D{.}{.}{2}|D{.}{.}{2}|D{.}{.}{2}|D{.}{.}{2}|D{.}{.}{2}|D{.}{.}{2}|}   
\hline
 & \textbf{en-da} & \textbf{en-is} & \textbf{en-no} & \textbf{da-is} & \textbf{da-no} & \textbf{is-no} \\
\hline
TDS & $0.58$ & $0.35$ & $0.60$ & $0.50$ & $0.92$ & $0.54$ \\ 
\hline
SYN & $0.50$ & $0.21$ & $0.41$ & $0.26$ & $0.42$ & $0.31$ \\ 
\hline
INV & $0.40$ & $0.40$ & $0.40$ & $1.00$ & $1.00$ & $1.00$ \\
\hline
PHON & $0.43$ & $0.43$ & $0.43$ & $0.99$ & $0.99$ & $0.99$ \\
\hline
\end{tabular}%
\end{adjustbox}
\end{table}

\begin{table}[t]
\centering
\caption{Percentage of examples per language in each language pre-training corpus, as classified by the fastText language identification model.}
\label{tab:language_contamination}
\vspace{1mm}
\begin{tabular}{|l|D{.}{.}{2}|D{.}{.}{2}|D{.}{.}{4}|D{.}{.}{2}|D{.}{.}{2}|}
\hline
\textbf{Language corpus} & \multicolumn{1}{c|}{\textbf{en}} & \multicolumn{1}{c|}{\textbf{da}} & \multicolumn{1}{c|}{\textbf{is}} & \multicolumn{1}{c|}{\textbf{no}} & \multicolumn{1}{c|}{\textbf{sv}}  \\
\hline
English & 99.79 & 0.06 & 0.0002 & 0.05 & 0.08  \\
\hline
Danish & 1.32 & 97.56 & 0.0003 & 0.98 & 0.12\\
\hline
Icelandic & 0.48 & 0.07 & 99.32 & 0.11 & 0.01 \\
\hline
 Norwegian & 3.16 & 1.12 & 0.001 & 95.40 & 0.24  \\
\hline
\end{tabular}%
\end{table}

\subsection{Language Contamination}
\label{sec:cont}
Another potentially important factor 
is language contamination, meaning that a language corpus may contain (small) amounts of text in other languages. The degree of contamination has been shown to correlate strongly with the performance of English monolingual models on other languages \cite{blevins-zettlemoyer-2022-language}.
To investigate such correlations in our 
scenario, we quantify the number of out-of-language raw text examples included in each language corpus, using the fastText language identification module \cite{joulin2016bag}. 

The language contamination analysis results are presented in Table \ref{tab:language_contamination}. 
English has the largest miscellaneous presence in the other corpora: around $3$\% of the Norwegian and around $1$\% of the Danish raw examples were classified as English. By contrast, the English training corpus contains only small amounts of data from the other languages, although Danish and Norwegian are more common than Icelandic. Icelandic, finally, is almost absent from the other training sets and also has the most homogeneous training set with less than 0.5\% of English and about 0.1\% each of Danish and Norwegian. We hypothesize that these asymmetries will affect the transfer effects between languages.
\begin{table}[t]
\caption{Test loss on English (en), Danish (da), Icelandic (is), and Norwegian (no), measured at the end of each pre-training sequence, for three model sizes.}
\label{tab:test_loss}
\vspace{1mm}
\centering
\begin{adjustbox}{width=0.85\textwidth}
\begin{tabular}{|l|r|r|r|r||r|r|r|r||r|r|r|r|}
\hline
\rule{0pt}{9pt}
&
\multicolumn{4}{|c||}{{\bf \small{Model: GPT 126M}}} &
\multicolumn{4}{|c||}{{\bf \small{Model: GPT 356M}}} &
\multicolumn{4}{|c|}{{\bf \small{Model: GPT 1.3B}}} \\
\hline
\textbf{} & \footnotesize{en} & \footnotesize{da} & \footnotesize{is} & \footnotesize{no} &
\footnotesize{en} & \footnotesize{da} & \footnotesize{is} & \footnotesize{no} & \footnotesize{en} & \footnotesize{da} & \footnotesize{is} & \footnotesize{no} \\
\hline
\rowcolor{Gray} 
\footnotesize{en}  & $3.45$ & $4.85$ & $6.38$ & $5.44$ & $3.08$ & $4.39$ & $5.99$ & $5.04$ & $2.79$ & $4.05$ & $5.70$ & $4.73$ \\
\footnotesize{da} & $3.34$ & $2.60$ & $5.82$ & $3.89$ & $3.20$ & $3.19$ & $4.97$ & $2.82$ & $2.92$ & $2.20$ & $5.43$ & $3.38$ \\
\rowcolor{Gray} 
\footnotesize{is} & $4.60$ & $5.25$ & $2.55$ & $5.79$ & $4.34$ & $4.95$ & $2.33$ & $5.50$ & $4.15$ & $4.73$ & $2.14$ & $5.27$ \\
\footnotesize{no} & $3.50$  & $3.47$ & $5.35$ & $3.11$ & $3.15$ & $2.41$ & $5.70$ & $3.65$ & $2.95$ & $2.95$ & $4.64$ & $2.57$ \\
\rowcolor{Gray} 
\footnotesize{en+da+is+no}  & $2.75$ & $2.84$ & $2.69$ & $3.29$ & $2.50$ & $2.56$ & $2.39$ & $2.97$ & $2.22$ & $2.26$ & $2.05$ & $2.61$ \\
\hline
\footnotesize{en-da} & $3.17$ & $2.54$ & - & - & $4.01$ & - & $2.23$ & - & $2.57$ & $2.03$ & - & - \\
\rowcolor{Gray} 
\footnotesize{en-is} & $4.43$ & - & $2.58$ & - & $3.02$ & - & - & $2.74$ & $3.89$ & - & $2.26$ & - \\
\footnotesize{en-no} & $3.41$ & - & - & $3.03$ & $2.84$ & $2.25$ & - & - & $2.73$ & - & - & $2.48$ \\
\hline
\rowcolor{Gray} 
\footnotesize{en-da-is} & $4.32$ & $4.84$ & $2.52$ & - & $3.91$ & $4.39$ & $2.19$ & - & $3.57$ & $3.91$ & $1.99$ & - \\
\footnotesize{en-da-no} & $3.35$ & $3.34$ & - & $3.04$ & $2.99$ & $2.99$ & - & $2.69$ & $2.72$ & $2.70$ & -  & $2.43$ \\
\rowcolor{Gray} 
\footnotesize{en-is-da} & $3.20$ & $2.52$ & $5.33$ & - & $2.86$ & $2.23$ & $4.88$ & - & $2.61$ & $2.01$ & $4.30$ & - \\
\footnotesize{en-is-no} & $3.42$ & - & $4.96$ & $3.07$ & $3.03$ & - & $4.48$ & $2.71$ & $2.82$ & - & $4.10$ & $2.51$ \\
\rowcolor{Gray} 
\footnotesize{en-no-da} & $3.16$ & $2.50$ & - & $3.65$ & $2.83$ & $2.22$ & -  & $3.26$ & $2.57$ & $1.99$ & - & $2.94$ \\
\footnotesize{en-no-is}  & $4.32$ & - & $2.54$ & $5.30$ & $3.91$ & - & $2.19$ & $4.92$ & $3.57$ & - & $1.97$ & $4.43$ \\
\hline
\rowcolor{Gray} 
\footnotesize{en-da-is-no} & $3.35$ & $3.36$ & $4.94$ & $3.03$ & $3.00$ & $3.01$ & $4.43$ & $2.68$ & $2.72$ & $3.83$ & $2.77$ & $2.43$ \\
\footnotesize{en-da-no-is }& $4.31$ & $4.85$ & $2.49$ & $5.37$ & $3.89$ & $4.36$ & $2.17$ & $4.89$ & $3.55$ & $3.92$ & $1.96$ & $4.39$ \\
\rowcolor{Gray} 
\footnotesize{en-is-da-no} & $3.35$ & $3.33$ & $5.12$ & $3.02$ & $2.98$ & $2.98$ & $4.68$ & $2.67$ & $2.70$ & $2.71$ & $4.13$ & $2.42$ \\
\footnotesize{en-is-no-da} & $3.16$ & $2.50$ & $5.43$ & $3.63$ & $2.84$ & $2.22$ & $5.01$ & $3.24$ & $2.62$ & $2.01$ & $4.54$ & $2.96$ \\
\rowcolor{Gray} 
\footnotesize{en-no-da-is} & $4.31$ & $4.83$ & $2.50$ & $5.41$ & $3.89$ & $4.34$ & $2.18$ & $4.94$ & $3.53$ & $3.86$ & $1.96$ & $4.45$ \\
\footnotesize{en-no-is-da} & $3.17$ & $2.50$ & $5.27$ & $3.70$ & $2.84$ & $2.21$ & $4.82$ & $3.31$ & $2.59$ & $1.99$ & $4.17$ & $3.00$\\
\hline
\end{tabular}%
\end{adjustbox}
\end{table}

\begin{figure}[t]
\centering
\includegraphics[width=\textwidth]{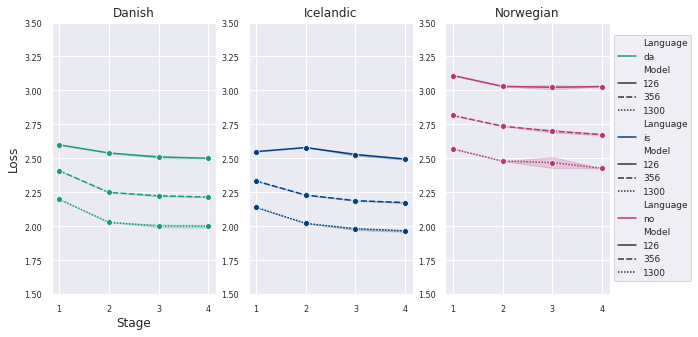}
\caption{Test loss on Danish, Icelandic, and Norwegian when 
learned at different stages. 
\emph{Clear improvement in the loss is observed when the language is learned later in the sequence, except for the $126M$ model trained on Icelandic.}
} 
\label{fig:forward}
\end{figure}

\section{Results}

Table \ref{tab:test_loss} shows 
test loss on 
all languages 
after each training stage. 
Starting with the baselines, the monolingual models 
in most cases reach a lower test loss than the corresponding jointly pre-trained model. The main exception is English, where joint pre-training leads to a lower test loss than 
monolingual training for all three model sizes. 
This is likely due to language contamination, as both the Danish and Norwegian training sets contain non-negligible amounts of English text. This hypothesis is supported
by the fact that monolingual Danish and Norwegian models attain competitive test loss on English. 

\subsection{Forward and Backward Transfer}
\label{sub:forward}

To visualize the forward transfer effects, Figure~\ref{fig:forward} shows the progression of model test loss on Danish, Icelandic and Norwegian when learned in the 2nd, 3rd or 4th training stage, in comparison to the monolingual baseline (1st stage). We observe a consistent positive trend for all languages, with diminishing returns as the length of the sequence training grows. The only exceptions are found for the $126$M model, where there is a slight loss increase for Icelandic in 2nd position and for Norwegian in 3rd position. This suggests that model capacity could be a limiting factor for positive forward transfer.

\begin{figure}[ht]
\centering
\subfloat{
  \includegraphics[width=.45\textwidth, height=3cm]{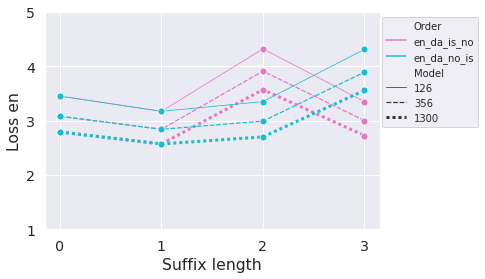}
}\hspace{.7em}
\subfloat{
  \includegraphics[width=.45\textwidth, height=3cm]{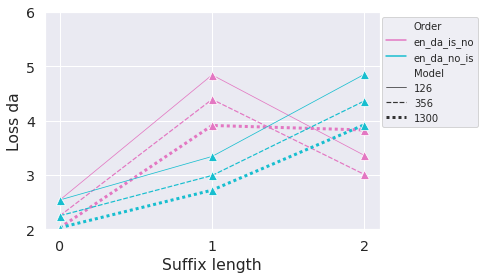}
}
\hfill
\subfloat{
  \includegraphics[width=.45\textwidth, height=3cm]{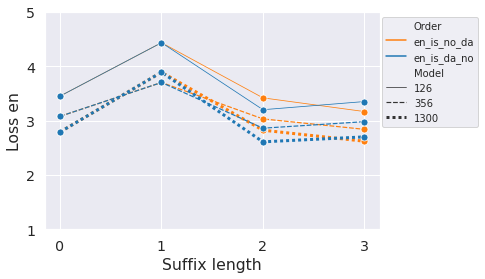}
}
\hspace{.7em}
\subfloat{
  \includegraphics[width=.45\textwidth, height=3cm]{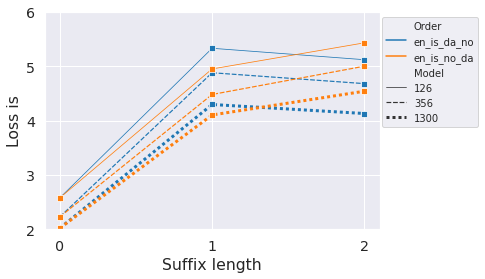}
}
\hfill
\subfloat{ 
  \includegraphics[width=.45\textwidth, height=3cm]{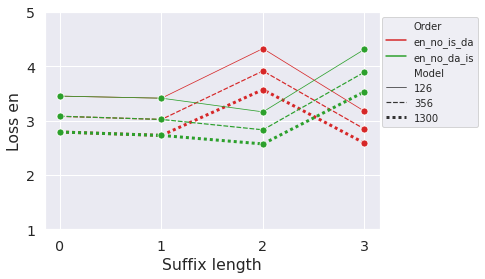}
}\hspace{.7em}
\subfloat{
  \includegraphics[width=.45\textwidth, height=3cm]{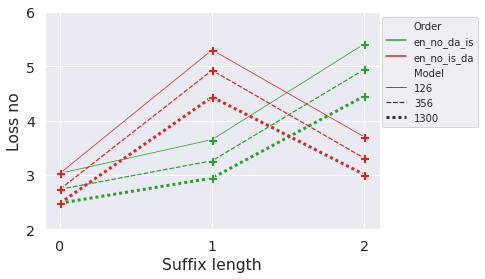}
}
\caption{Suffix length refers to the number of languages added after then one visualised. Test loss on English (\textbf{Left}) and on Danish, Icelandic, and Norwegian (\textbf{Right}) for models with varying size and language suffixes. \emph{Overall, Icelandic always causes forgetting to the other languages, while positive (or weaker negative) transfer is observed between Danish and Norwegian, and from those two languages to English.}}
\label{fig:backward}
\end{figure}

The research question about backward transfer concerns what happens to previously seen languages when a new language is learned. The left-hand side of Figure~\ref{fig:backward} visualizes the evolution of model test loss on English, which is always the first training language, as other languages are added in different order. The patterns here are strikingly different from the ones for forward transfer in that they do not relate to the position of a language in the sequence, but to the choice of language. More specifically, training a model on Icelandic always leads to an \emph{increase} in the English test loss, which is a sign of catastrophic forgetting or negative transfer. By contrast, training a model on Danish always leads to a \emph{decrease} in the English test loss, regardless of where it occurs in the sequence. Training on Norwegian, finally, has an effect in between the two extremes; it mostly leads to a decrease in English test loss, although smaller than for Danish, but it leads to an increase when added directly after Danish. 

The right-hand side of Figure~\ref{fig:backward} instead visualizes the test loss for Danish, Icelandic, and Norwegian, respectively, as the other two languages are added in different order. In this case, Icelandic still hurts the other two and also suffers a degradation in performance when one of the other two is added. By contrast, Danish and Norwegian have a much weaker negative effect on each other when adjacent, and mutually help each other when added after Icelandic.
The overall pattern emerging from these observations about negative transfer 
is that Icelandic interacts negatively with all the other languages, while Danish and Norwegian mostly have a weak positive effect on English and each other, partly dependent on their position in the training sequence.

\subsection{Model Size}
\label{sub:size}
\begin{figure}[t]
\centering
{
 \includegraphics[width=0.45\textwidth, height=3.8cm]{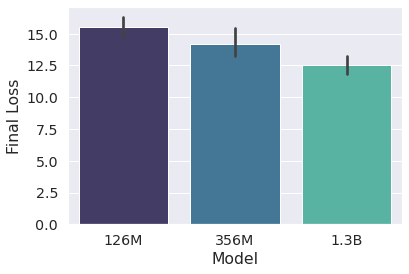}
}\hspace{.7em}
{
    \includegraphics[width=0.45\textwidth, height=3.8cm]{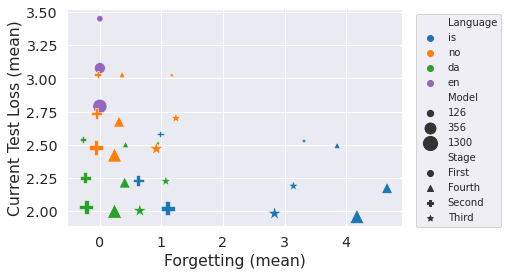}
}
\caption{\textbf{Left:} Cumulative loss over all language test sets, averaged per model size at the final stage. 
\emph{Growing the model size from $126$M to $356$M and then to $1.3$B leads to a drop of the model test loss on average by $8$\% and $11$\% respectively}.
\textbf{Right:} Test loss in the current language vs forgetting ( i.e. loss growth on previous languages ). \emph{For a given language and stage (color and shape), increasing the size of the model consistently decreases the current loss while forgetting 
remains mostly the same.}} 
 \label{fig:trade-off}
\end{figure}

Our final research question concerns the effect of model size on forward and backward transfer. As shown in Figure~\ref{fig:trade-off} (left), overall test loss decreases as expected with increasing model size. However, contrary to our expectations, the forgetting patterns do not change noticeably when increasing the model size. This is reflected already in the largely parallel curves in Figures~\ref{fig:backward}, but is visualized directly in Figure~\ref{fig:trade-off} (right), which shows the trade-off between test loss in the current language and the cumulative loss increase on previous languages. Symbols representing different model sizes for the same language and sequence position are vertically aligned in most cases, indicating that they achieve different loss levels with a constant negative backward transfer effect, except for the largest model when the target language is Icelandic.
\section{Discussion}
\label{sec:discussion}
One of our main results is that, while positive forward transfer appears to be language-independent, 
backward transfer is sensitive to cross-linguistic interaction. 
A concept often invoked in this context is language similarity \cite{philippy-etal-2023-towards}, where the idea is that similar languages support each other, while dissimilar languages may harm each other. Section~\ref{sec:sim} reports on four different similarity metrics, three derived from the URIEL database \cite{littell-etal-2017-uriel} and one novel metric based on token distribution similarity (TDS). 

Starting with the URIEL metrics, we note that the phonetic (INV) and phonological (PHON) 
do not appear to be informative; they both show a very high similarity between Danish, Icelandic and Norwegian, and a lower similarity between these languages and English, which is incompatible with the observed patterns. 
Syntactic similarity (SYN) looks more promising, as it ranks the language pairs in a way that is compatible with the observed interactions: en-da $>$ da-no $>$ en-no $>$is-no $>$ da-is $>$ en-is. The TDS metric generally shows the lowest values for pairs involving Icelandic, which is compatible with the observed negative backward transfer effects for this language. However, only the value for Icelandic and English is significantly lower than for other language pairs. 
Moreover, despite a very high similarity between Danish and Norwegian, we do not observe a strong positive backward transfer between these languages, so the explanatory value of the TDS metric seems limited. 

For language contamination, \cite{blevins-zettlemoyer-2022-language} shows that the performance of (supposedly) monolingual models for English on other languages is strongly correlated with the amount of data (accidentally) included from the respective languages. The language contamination statistics in Section~\ref{sec:cont} 
clearly show that Icelandic is extreme both in having less contamination in its training set (especially for Danish and Norwegian) and in having negligible amounts of data in the other training sets, which suggests that this is the primary cause of the strong negative transfer effects involving Icelandic. Language contamination could also explain why Danish and Norwegian have stronger positive effects on English than on each other, given that these training sets contain the largest amount of English data among the non-English training sets. However, 
it does not explain why Danish has a stronger positive effect than Norwegian, since the amount of English data is more than twice as large in the Norwegian training set. The fact that Danish shows a larger syntactic similarity to English than Norwegian suggests that we need to take both contamination and linguistic similarity into account. 

\section{Conclusion}

In this paper, we study forward and backward transfer in a language modeling task -- standardly used for pre-training of large language models -- in a novel multilingual continual learning scenario. 
Our experimental results with three model sizes 
and four languages 
indicate that forward transfer is consistently positive. However, backward transfer can be positive or negative, depending on the choice of languages and their order.
In an attempt to further analyze the transfer patterns observed, we 
find that no single factor can explain all patterns and that a combination of language contamination and syntactic similarity 
best fits our empirical results. Increasing model size improves performance for all languages, but 
all transfer patterns remain essentially the same.

We hope that our findings will help design large-scale continual learning language models 
The next step would be to explore a larger and more diverse set of languages to assess the generality of the transfer patterns. As a long-term goal, this could improve resource efficiency while reducing negative transfer in the computationally extensive setup of continual language model pre-training.

\paragraph{\textbf{Acknowledgements}}
The research presented in this paper was supported by the Swedish Research Council (grant no. 2022-02909).
A significant part of the computations was enabled by the Berzelius resource provided by the Knut and Alice Wallenberg Foundation at the National Supercomputer Centre at Linköping University, Sweden (Berzelius-2023-178).
In addition, the authors gratefully acknowledge the HPC RIVR consortium (www.hpc-rivr.si) and EuroHPC JU (eurohpc-ju.europa.eu) for funding this research by providing computing resources of the HPC system Vega at the Institute of Information Science (www.izum.si). 
Magnus Boman acknowledges funding from the Swedish Research Council on Scalable Federated Architectures.

\bibliographystyle{splncs04}
\bibliography{custom,continualJN}

\appendix
\section{Experimental Details}
\label{app:details}

Table \ref{tab:model_hyperpar} shows parameters for each model size.  
Table \ref{tab:data_weights_normalised} shows 
dataset weights. 

\vspace*{-8mm}
\begin{table}[h!]
\caption{Model and training parameters for the three model sizes.}
\label{tab:model_hyperpar}
\centering
{\renewcommand{\arraystretch}{0.9}%
\begin{tabular}{|l|c|c|c|}
\hline
 \textbf{Parameter Name} & \textbf{126M Model} & \textbf{356M Model} & \textbf{1.3B Model}\\
 \hline
Sequence length & $2048$ & $2048$ & $2048$\\
 Number of layers & $12$ & $24$ & $24$\\
 Hidden size dimension & $768$ &$1024$ & $2048$\\
 Number of attention heads & $12$ & $16$ & $32$\\
\hline
Tensor parallelism & $1$ & $2$ & $4$\\
Micro batch size & $4$ & $4$ & $4$\\
Global batch size & $1024$ & $1024$ & $1024$\\
 Min learning rate & $6e-5$ & $3e-5$ & $2e-5$\\
 Max learning rate & $6e-4$ & $3e-4$ & $2e-4$\\
 \hline
\end{tabular}
}

{\renewcommand{\arraystretch}{0.9}%
\caption{Datasets with size and assigned weighted contribution to the language corpus.}\label{tab:data_weights_normalised}
\centering
\begin{tabular}{|l|c|c|c|c|}
\hline
\textbf{Dataset} & \textbf{Category} & \textbf{Weight} & \textbf{Dataset size (GB)} & \textbf{Normalised weight} \\
\hline
books3 (The Pile) &  Books & $1$ & $89$ & $0.43$ \\
PubMed, arXiv (The Pile)  &  Articles & $0.9$ & $33$ & $0.15$\\
Stackexchange (The Pile) & Miscellaneous & $1$ & $35$ & $0.17$\\
NCC & Miscellaneous & $1$ & $39$ & $0.49$\\
Icelandic Gigaword & Miscellaneous & $1$ & $9.6$ & $0.69$ \\
Danish Gigaword & Miscellaneous & $1$ & $3.47$ & $0.06$ \\
Pile Openwebtext & Web CC & $0.5$ & $58$ & $0.14$\\
mc4, oscar (no)  & Web CC & $0.5$ & $78$ & $0.49$\\
mc4, oscar (is) & Web CC & $0.5$ & $8.4$ & $0.34$\\
mc4, oscar (da) & Web CC & $0.5$ & $93$ & $0.93$\\
Wiki en & Wiki & $1.5$ & $15$ & $0.1$\\
Wiki no & Wiki & $1.5$ & $0.5$ & $0.01$\\
Wiki is & Wiki & $1.5$ & $0.05$ & $0.006$\\
Wiki da & Wiki & $1.5$ & $0.4$ & $0.01$\\

\hline

\end{tabular}
}
\end{table}
\vspace{-5mm}

\noindent
Our experiments ran in two clusters
: Vega \footnote{\url{https://doc.vega.izum.si/}} a GPU cluster part of the EuroHPC network consisting of 60 GPU nodes of 4 NVidia A100 each and Berzelius \footnote{\url{https://www.nsc.liu.se/systems/berzelius/}} which is located in Linköping, Sweden and consists of 94 GPU nodes of 8 NVidia A100 each. A single monolingual training phase of the $126$M model required $352$ GPU hours on Vega, while the same run for the $356$M model took $352$ GPU hours on Berzelius. All $1.3$B models were trained on Berzelius and each of the pre-training stages on a single language took $2272$ GPU hours.

\end{document}